%% file: main.tex
\documentclass[conference]{IEEEtran}
\IEEEoverridecommandlockouts

\usepackage{cite}
\usepackage{amsmath,amssymb,amsfonts}
\usepackage{algorithmic}
\usepackage{graphicx}
\usepackage{textcomp}
\usepackage{xcolor}
\usepackage{caption}
\usepackage{subcaption}
\usepackage{booktabs}
\usepackage{multirow}
\usepackage{makecell}
\usepackage{array}
\def\BibTeX{{\rm B\kern-.05em{\sc i\kern-.025em b}\kern-.08em
    T\kern-.1667em\lower.7ex\hbox{E}\kern-.125emX}}

\begin{document}
\bstctlcite{IEEEexample:BSTcontrol}
\title{MP-MoE: Matrix Profile-Guided Mixture of Experts for Precipitation Forecasting
}


\author{
  \IEEEauthorblockN{
    Huyen Ngoc Tran\IEEEauthorrefmark{2},
    Dung Trung Tran\IEEEauthorrefmark{2},
    Hong Nguyen\IEEEauthorrefmark{3},
    Xuan Vu Phan\IEEEauthorrefmark{2},
    Nam-Phong Nguyen\IEEEauthorrefmark{2}
  }
  \IEEEauthorblockA{
    \IEEEauthorrefmark{2}School of Electronics and Electrical Engineering, Hanoi University of Science and Technology
  }
  \IEEEauthorblockA{
    \IEEEauthorrefmark{3}Department of Electrical and Computer Engineering,
    University of Southern California, United States
  }
}

\maketitle
\begin{abstract}
Precipitation forecasting remains a persistent challenge in tropical regions like Vietnam, where complex topography and convective instability often limit the accuracy of Numerical Weather Prediction (NWP) models. While data-driven post-processing is widely used to mitigate these biases, most existing frameworks rely on point-wise objective functions, which suffer from the ``double penalty'' effect under minor temporal misalignments. In this work, we propose the Matrix Profile-guided Mixture of Experts (MP-MoE), a framework that integrates conventional intensity loss with a structural-aware Matrix Profile objective. By leveraging subsequence-level similarity rather than point-wise errors, the proposed loss facilitates more reliable expert selection and mitigates excessive penalization caused by phase shifts. We evaluate MP-MoE on rainfall datasets from two major river basins in Vietnam across multiple horizons, including 1-hour intensity and accumulated rainfall over 12, 24, and 48 hours. Experimental results demonstrate that MP-MoE outperforms raw NWP and baseline learning methods in terms of Mean Critical Success Index (CSI-M) for heavy rainfall events, while significantly reducing Dynamic Time Warping (DTW) values. These findings highlight the framework's efficacy in capturing peak rainfall intensities and preserving the morphological integrity of storm events.
\end{abstract}

\begin{IEEEkeywords}
Double penalty, Matrix Profile, Mixture of Experts, NWP post-processing, Precipitation Forecasting.
\end{IEEEkeywords}
\input{secs/introduction}
\input{secs/related_work}
\input{secs/methodology}
\input{secs/Experiments_Results}
\input{secs/Limitations}
\input{secs/Conclusion}
\bibliographystyle{IEEEtran}
\bibliography{references}
\end{document}

%% file: secs/introduction.tex
\section{Introduction}
\label{sec:intro}
   Precipitation forecasting plays a vital role in mitigating natural hazards, optimizing reservoir operations, and ensuring sustainable agricultural planning. Traditionally, Numerical Weather Prediction (NWP) models~\cite{Bauer2015nwp,fritsch1998quantitative,mitsuishi2011applicability} have been globally adopted as the standard for short-term rainfall forecasting due to their ability to simulate atmospheric dynamics. Various operational frameworks, such as the Weather Research and Forecasting (WRF) system~\cite{skamarock2008wrf}, have served as the primary physical foundation for these efforts by solving complex fluid dynamics and thermodynamic equations under specific initial and boundary conditions. However, despite their widespread applications, they are inherently constrained by the chaotic nature of the atmosphere and their sensitivity to initial perturbations. Such limitations often manifest as systematic biases and spatial displacement errors, particularly within tropical regions characterized by complex topography, such as Vietnam.
    
    Recently, statistical post-processing has emerged as a well-established framework to calibrate raw model outputs, encompassing a wide range of methodologies. These approaches span from classic probabilistic techniques such as Bayesian Model Averaging (BMA)~\cite{raftery2005using} and Quantile Regression Averaging (QRA)~\cite{bremnes2004probabilistic} to more recent advanced data-driven models including XGBoost~\cite{chen2016xgboost} and Long Short-Term Memory (LSTM)~\cite{Hochreiter1997lstm}. While these methods have successfully improved calibration for general weather conditions and captured complex nonlinear dependencies, they often struggle to accurately resolve rapidly evolving convective systems or extreme rainfall onsets. One of the fundamental limitations is that they rely on pointwise loss functions like Mean Squared Error (MSE) or Mean Absolute Error (MAE). This dependency triggers the well-known ``double penalty'' problem, where a forecast that correctly captures a storm's structural evolution but is slightly displaced in time is penalized twice: once for the displacement and once for the perceived false alarm~\cite{ebert2008fuzzy,Wang2009}. Mathematically, such an optimization landscape incentivizes models to produce overly smoothed, blurred predictions to minimize variance, resulting in the systematic suppression of peak rainfall intensities, a phenomenon commonly referred to as ``peak-shaving''~\cite{ravuri2021skilful}.

    To bridge these critical gaps, we propose the Magnitude-Aware Matrix Profile Mixture of Experts (MP-MoE). This neural post-processing framework shifts the optimization paradigm from strict point-wise accuracy to structural fidelity by integrating a shape-aware objective into a dynamic expert selection mechanism. Unlike traditional methods, MP-MoE leverages temporal subsequence similarity to mitigate the double penalty effect while preserving the physical magnitude of extreme events. The key contributions of this work are summarized as follows:
    \begin{itemize}
        \item A hybrid loss function is introduced that integrates MSE with a structural similarity penalty derived from an unnormalized MP distance~\cite{yeh2016matrix}. This mechanism acts as a ``soft'' time-lag correction, rewarding models that capture the true physical evolution of rainfall.
        \item A gating network is proposed that uses large-scale GFS variables to dynamically assign weights to specific WRF experts based on meteorological regimes. This architecture allows the system to adaptively switch strategies between stable stratiform and convective rainfall regimes.
        \item Comprehensive experiments are conducted in two real-world hydrological basins in Vietnam, demonstrating that MP-MoE significantly reduces Dynamic Time Warping (DTW) values~\cite{keogh2001derivative} and reconstructs peak intensities.
    \end{itemize}

%% file: secs/related_work.tex
\section{Related Work}
\label{sec:related_work}
    Rainfall forecasting has traditionally relied on NWP models, which provide a rigorous physical foundation but frequently exhibit systematic biases due to simplified atmospheric parameterizations and coarse spatial resolutions~\cite{Bauer2015nwp,vannitsem2021statistical}. To mitigate these discrepancies, statistical post-processing methods, such as BMA and QRA, have been widely implemented to refine raw NWP outputs. While effective in improving probabilistic calibration, these frameworks often struggle to capture the complex, non-linear dependencies inherent in extreme meteorological events, particularly during rapid storm onsets where atmospheric conditions shift abruptly~\cite{Graefe2015bma,vannitsem2021statistical}. 
    
    Recent advances in deep learning have accelerated the shift toward data-driven rainfall forecasting models. Early architectures advanced data-driven forecasting by transitioning from temporal modeling with LSTM to capturing joint spatiotemporal dependencies via ConvLSTM and PredRNN~\cite{shi2015convlstm,wang2022predrnn}. These foundations paved the way for high-capacity frameworks such as GraphCast~\cite{lam2023graphcast}, Earthformer~\cite{gao2023earthformer}, and Pangu-Weather~\cite{bi2022panguweather}, which excel at extracting multi-scale patterns from atmospheric data. Despite their success, a persistent challenge is that many such models rely predominantly on point-wise objective functions like MSE. For highly non-stationary rainfall processes, these losses tend to penalize temporal misalignment excessively, leading to the well-known ``peak-shaving'' effect where intense precipitation events are smoothed or underestimated~\cite{ebert2008fuzzy}. To address the limitations of point-wise accuracy, structural distance measures such as DTW and Soft-DTW have been proposed to alleviate this issue by allowing elastic temporal matching between predicted and observed rainfall sequences~\cite{Salvador2007,Cuturi2017softdtw}. Nevertheless, these methods suffer from quadratic computational complexity and may introduce non-physical temporal warpings, limiting their practicality for operational hydrological forecasting. While the MP paradigm has recently emerged as an efficient framework for subsequence similarity search in data mining~\cite{yeh2016matrix, tran2025mp}, its potential as a structural penalty mechanism to guide expert selection remains largely unexploited. Unlike DTW’s unrestricted temporal warping, MP identifies structurally similar patterns within a constrained temporal search range, thereby preserving the morphological characteristics of rainfall events while maintaining a much lower constant-factor overhead in practical execution.
    
    In parallel with loss function optimization, the MoE architecture has received increasing attention for managing heterogeneous data regimes. Unlike static ensembles that apply fixed weights to model outputs, MoE utilizes a dynamic gating mechanism to route inputs to specialized experts based on the current context~\cite{jacobs1991adaptive,shazeer2017outrageously}. While MoE has shown promise in general time-series tasks, its application in NWP post-processing remains under-explored, particularly regarding the use of structural similarity to guide the gating process. Our approach bridges this gap by integrating MP-derived guidance into an MP-MoE architecture. By optimizing the selection process through a redefined structural loss, our framework ensures adaptive expert selection while preserving the integrity of rainfall hydrographs, effectively connecting large-scale physical modeling with localized structural accuracy.

%% file: secs/methodology.tex
\section{Methodology}
\label{sec:Method}
    \begin{figure*}[t!]
        \centering
        \includegraphics[width=0.98\textwidth]{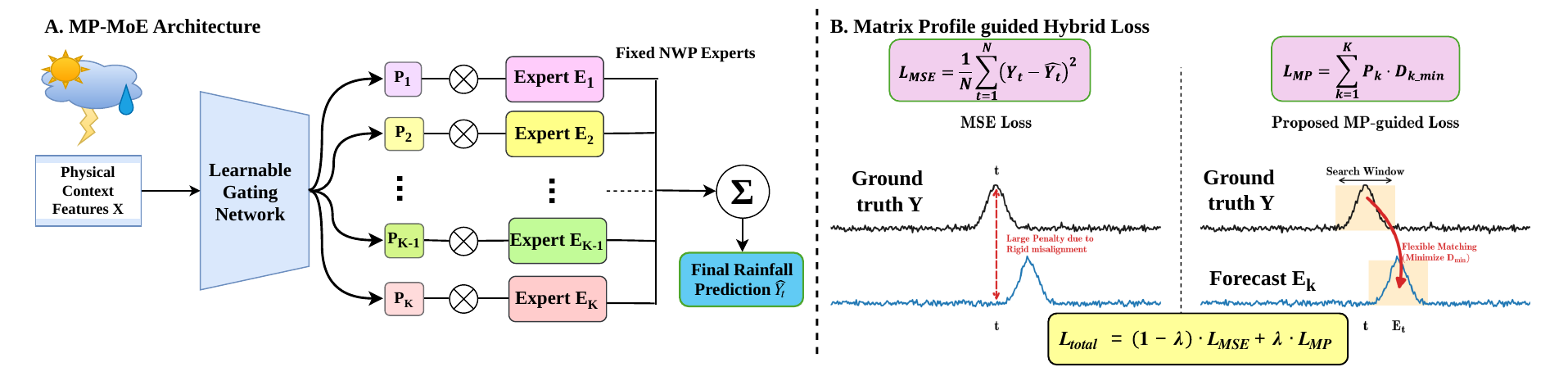}
        \caption{Schematic overview of the proposed MP-MoE framework. Panel A (left) illustrates the inference workflow where the learnable gating network processes large-scale physical features to assign importance weights to fixed NWP experts dynamically. Panel B (right) details the training strategy using a hybrid loss function. Unlike standard MSE, which imposes severe penalties for temporal shifts, the proposed matrix profile-guided objective scans a symmetric search window to minimize structural distance, thereby prioritizing shape fidelity over rigid alignment.}
        \label{fig:method_overview}
    \end{figure*}
    \subsection{Data Preprocessing and Feature Engineering}
    To ensure numerical stability, all physical features are Z-score normalized. To facilitate the MP-guided objective, we structure the data into two distinct temporal sequences: (i) an expert query sequence, comprising a 3-hour forecast trajectory from $t-2$ to $t$; and (ii) a ground truth search scope, spanning a symmetric 6-hour window centered at the target time, from $t-3$ to $t+3$. By using the expert sequence to query this extended symmetric window, the loss function can identify the optimal structural match regardless of whether the forecast arrives earlier or later than observed. Consequently, samples at the boundaries of the timeline, which lack sufficient data to form the complete symmetric search window $[t-\Delta, t+\Delta]$, were excluded from the training set.
    \subsection{Proposed Framework: Matrix Profile-Guided MoE}

    The core of the MP-MoE framework is a gating network $G$, which dynamically determines the reliability of each expert based on the local meteorological state. At each time step $t$, the network processes a vector of large-scale physical features $X_t$ (e.g., humidity, wind components from GFS) to modulate the contribution of individual NWP experts. The gating mechanism, projects these features into a latent representation, followed by a Softmax activation to generate a probability distribution over the $K$ experts:
    \begin{equation}
        p_{k,t} = \frac{\exp(G_k(X_t))}{\sum_{j=1}^K \exp(G_j(X_t))},
    \end{equation}
    where $G_k(X_t)$ represents the logit assigned to the k-th expert, and $p_{k,t}$ denotes the adaptive weight assigned to the $k-th$ expert, satisfying $\sum_{k=1}^Kp_{k,t}=1$ and $p_{k,t}\ge0$.
    The final rainfall estimate $\hat{Y_t}$ is then computed as a weighted linear combination:
    \begin{equation}
        \hat{Y}t = \sum_{k=1}^{K} p_{k,t} \cdot E_{k,t},
    \end{equation}
    where $E_{k,t}$ is the forecast from the $k-th$ NWP expert. 
    \subsection{Hybrid Objective Function}
    Unlike canonical MP, which relies on Z-normalization to find shape motifs regardless of amplitude, we introduce an unnormalized Euclidean distance to ensure mass conservation and penalize volumetric errors. Although MSE loss functions are crucial for calibrating the overall rainfall magnitude, sole reliance on them yields smoothed predictions due to the ``double penalty'' effect discussed in section \ref{sec:related_work}. Therefore, we propose a hybrid objective function that balances two complementary goals: (i) minimizing the rainfall intensity error via MSE, and (ii) preserving the temporal shape of storm events via an MP-based guidance.
    \begin{equation}
        \mathcal{L}_{\text{total}} = (1-\lambda)\cdot\mathcal{L}_{\text{MSE}} + \lambda\cdot\mathcal{L}_{\text{MP}}.
        \label{eq:loss_total}
    \end{equation}
    \noindent\textbf{Rainfall intensity calibration.} The first component of \eqref{eq:loss_total} focuses on minimizing the error in rainfall magnitude. We employ the standard MSE between the aggregated forecast $\hat{Y}$ and the ground truth $Y$. 
    \begin{equation}
        \mathcal{L}_{\text{MSE}} = \frac{1}{N}\sum_{i=1}^{N}(Y_i-\hat{Y_i})^2.
    \end{equation}
    \noindent\textbf{Temporal shape preservation.} The second component of \eqref{eq:loss_total} penalizes structural dissimilarity rather than point-wise displacement. We use the MP distance to align the temporal peaks of the forecast with the observations:
    \begin{equation}
    \mathcal{L}_{\text{MP}} = \frac{1}{N}\sum_{t=1}^{N}\sum_{k=1}^K P_{k,t}\cdot D_{\text{min}}(\mathbf{E}^{(t)}_k,\mathbf{Y}^{(t)}),
    \end{equation}
    where $P_{k,t}$ denotes the probability weight assigned to the $k$-th NWP expert for the $t$-th sample, while $N$ and $K$ are the number of training samples and experts, respectively. The term $D_{\min}(\mathbf{E}^{(t)}_k, \textbf{Y}^{(t)})$ is the MP distance (visualized in Fig.~\ref{fig:method_overview}), calculated as the minimum Euclidean distance between the sliding window of the expert forecast and its best matching subsequence in the ground truth:
    \begin{equation}
    D_{\min}(\mathbf{E}_k, \mathbf{Y}) = \min_{\tau \in [t-\Delta, t+\Delta]} \| \mathbf{w}^{\text{E}}_{k,t} - \mathbf{w}^{\text{Y}}_\tau \|_2,
    \end{equation}
    where the sliding windows of length $m$ are defined as $\mathbf{w}^{\text{E}}_{k,t} = [e_{k, t-m+1}, \dots, e_{k, t}]$ for the $k$-th expert ending at time $t$, and $\mathbf{w}^{\text{Y}}_\tau = [y_{\tau-m+1}, \dots, y_\tau]$ for the candidate ground truth window ending at $\tau$. The parameter $\Delta$ represents the maximum permissible temporal shift, creating a symmetric search scope $[t-\Delta, t+\Delta]$. The operation $\min_\tau$ searches within this scope to effectively handle both early and delayed arrival of rainfall events. Crucially, since the NWP expert forecasts $E$ and ground truth $Y$ are fixed inputs, the term $D_{min}$ functions as a static penalty coefficient. This approach effectively simplifies the optimization process, as the Gating Network can update its parameters based on pre-computed modified MP distances without the need to differentiate through discrete time indices.

%% file: secs/Experiments_Results.tex
\section{Experiments and Results}
\label{sec:experiments_results}
    \subsection{Experimental Settings}
    \noindent\textbf{Datasets.}~The proposed framework is evaluated using real-world rainfall data from the Ban Nhung and Song Chay basins, two mountainous hydrological catchments located in northern Vietnam. These regions are characterized by steep terrain with elevations ranging from low valleys to over 2,400 m, resulting in strong elevation gradients and pronounced orographic influences on rainfall. The Song Chay basin contains a dense river network and receives high annual precipitation, typically exceeding 2,000 mm. In contrast, the Ban Nhung basin frequently experiences intense localized convective rainfall events associated with complex mountainous meteorology. Such environmental conditions often lead to rapid runoff and flash flooding during heavy rainfall episodes, making these basins suitable testbeds for evaluating a model’s ability to capture both spatial displacement and peak rainfall intensity. We employ a 7:3 chronological split to test the model across diverse meteorological regimes and terrain-induced variability.

    \begin{table*}[t!]
        \centering
        \caption{Quantitative performance comparison on the Ban Nhung and Song Chay basins. Evaluation metrics include MAE for 1-hour intensity and accumulated rainfall (12h, 24h, 48h), DTW distance for structural similarity, and CSI-M for event detection. \textit{Note:} For learning-based models, results are reported as Mean $\pm$ STD over five independent runs; physics-based experts are deterministic and report a single values. MP-MoE ($\lambda=0.6$) denotes our proposed method, showing consistent improvements over both raw WRF experts and learning-based methods. \textbf{Bold} values indicate the best performance for each metric, while \underline{underlined values} denote the second best.}
        \label{tab:results_table}
    \resizebox{\textwidth}{!}{    
    \begin{tabular}{|llcccccc|}
    \toprule
    \multicolumn{2}{|l|}{\multirow{2}{*}{Basin / Method}} & \multicolumn{4}{c|}{\textbf{MAE [mm] (Accumulated Rainfall) $\downarrow$}} & \multicolumn{1}{c|}{\multirow{2}{*}{\textbf{DTW $\downarrow$}}} & \multirow{2}{*}{\textbf{CSI-M $\uparrow$}} \\ \cmidrule(lr){3-6}
    \multicolumn{2}{|l|}{} & \multicolumn{1}{c|}{\textbf{1h}} & \multicolumn{1}{c|}{\textbf{12h}} & \multicolumn{1}{c|}{\textbf{24h}} & \multicolumn{1}{c|}{\textbf{48h}} & \multicolumn{1}{c|}{} & \\ \midrule \midrule
    \multicolumn{8}{|c|}{\textbf{Panel A: Ban Nhung Basin}} \\ \midrule
    \multicolumn{1}{|l|}{\multirow{9}{*}{\begin{tabular}[c]{@{}l@{}}Physics-Based\\ Experts\end{tabular}}} & \multicolumn{1}{l|}{COMS} & \multicolumn{1}{c|}{0.669} & \multicolumn{1}{c|}{7.243} & \multicolumn{1}{c|}{14.246} & \multicolumn{1}{c|}{27.739} & \multicolumn{1}{c|}{2074.0} & 0.068 \\
    \multicolumn{1}{|l|}{} & \multicolumn{1}{l|}{GFS} & \multicolumn{1}{c|}{0.689} & \multicolumn{1}{c|}{7.536} & \multicolumn{1}{c|}{14.948} & \multicolumn{1}{c|}{29.338} & \multicolumn{1}{c|}{1985.9} & 0.041 \\
    \multicolumn{1}{|l|}{} & \multicolumn{1}{l|}{MIT\_D02} & \multicolumn{1}{c|}{1.000} & \multicolumn{1}{c|}{10.930} & \multicolumn{1}{c|}{21.536} & \multicolumn{1}{c|}{41.996} & \multicolumn{1}{c|}{2936.9} & 0.112 \\
    \multicolumn{1}{|l|}{} & \multicolumn{1}{l|}{LING3\_D02} & \multicolumn{1}{c|}{0.850} & \multicolumn{1}{c|}{9.042} & \multicolumn{1}{c|}{17.817} & \multicolumn{1}{c|}{34.733} & \multicolumn{1}{c|}{2684.3} & 0.096 \\
    \multicolumn{1}{|l|}{} & \multicolumn{1}{l|}{LINKF\_D02} & \multicolumn{1}{c|}{1.014} & \multicolumn{1}{c|}{10.953} & \multicolumn{1}{c|}{21.511} & \multicolumn{1}{c|}{41.750} & \multicolumn{1}{c|}{3007.4} & 0.109 \\
    \multicolumn{1}{|l|}{} & \multicolumn{1}{l|}{LINBMJ\_D02} & \multicolumn{1}{c|}{0.692} & \multicolumn{1}{c|}{6.534} & \multicolumn{1}{c|}{12.656} & \multicolumn{1}{c|}{24.086} & \multicolumn{1}{c|}{2339.0} & 0.113 \\
    \multicolumn{1}{|l|}{} & \multicolumn{1}{l|}{ETAKF\_D02} & \multicolumn{1}{c|}{0.859} & \multicolumn{1}{c|}{9.104} & \multicolumn{1}{c|}{17.813} & \multicolumn{1}{c|}{34.730} & \multicolumn{1}{c|}{2609.6} & 0.116 \\
    \multicolumn{1}{|l|}{} & \multicolumn{1}{l|}{ETAG3\_D02} & \multicolumn{1}{c|}{0.778} & \multicolumn{1}{c|}{8.264} & \multicolumn{1}{c|}{16.252} & \multicolumn{1}{c|}{31.744} & \multicolumn{1}{c|}{2257.2} & 0.106 \\
    \multicolumn{1}{|l|}{} & \multicolumn{1}{l|}{ETABMJ\_D02} & \multicolumn{1}{c|}{0.592} & \multicolumn{1}{c|}{5.527} & \multicolumn{1}{c|}{10.641} & \multicolumn{1}{c|}{20.078} & \multicolumn{1}{c|}{1928.4} & \underline{0.121} \\ \midrule
    \multicolumn{1}{|l|}{\multirow{9}{*}{\begin{tabular}[c]{@{}l@{}}Learning-based\\ Models\end{tabular}}} & \multicolumn{1}{l|}{Ensemble} & \multicolumn{1}{c|}{0.546 $\pm$ ${\scriptstyle 0.007}$} & \multicolumn{1}{c|}{6.363 $\pm$ ${\scriptstyle 0.074}$} & \multicolumn{1}{c|}{12.530 $\pm$ ${\scriptstyle 0.139}$} & \multicolumn{1}{c|}{24.601 $\pm$ ${\scriptstyle 0.266}$} & \multicolumn{1}{c|}{1626.4 $\pm$ ${\scriptstyle 47.0}$} & 0.004 $\pm$ ${\scriptstyle 0.004}$ \\
    \multicolumn{1}{|l|}{} & \multicolumn{1}{l|}{XGBoost} & \multicolumn{1}{c|}{0.622 $\pm$ ${\scriptstyle 0.006}$} & \multicolumn{1}{c|}{7.092 $\pm$ ${\scriptstyle 0.082}$} & \multicolumn{1}{c|}{13.926 $\pm$ ${\scriptstyle 0.164}$} & \multicolumn{1}{c|}{27.110 $\pm$ ${\scriptstyle 0.312}$} & \multicolumn{1}{c|}{1325.8 $\pm$ ${\scriptstyle 38.4}$} & 0.080 $\pm$ ${\scriptstyle 0.002}$ \\
    \multicolumn{1}{|l|}{} & \multicolumn{1}{l|}{LSTM} & \multicolumn{1}{c|}{0.584 $\pm$ ${\scriptstyle 0.024}$} & \multicolumn{1}{c|}{6.026 $\pm$ ${\scriptstyle 0.235}$} & \multicolumn{1}{c|}{11.836 $\pm$ ${\scriptstyle 0.483}$} & \multicolumn{1}{c|}{23.271 $\pm$ ${\scriptstyle 0.982}$} & \multicolumn{1}{c|}{\underline{997.0 $\pm$ ${\scriptstyle 51.5}$}} & 0.024 $\pm$ ${\scriptstyle 0.003}$ \\
    \multicolumn{1}{|l|}{} & \multicolumn{1}{l|}{QRA} & \multicolumn{1}{c|}{\underline{0.440 $\pm$ ${\scriptstyle 0.005}$}} & \multicolumn{1}{c|}{\underline{5.065 $\pm$ ${\scriptstyle 0.062}$}} & \multicolumn{1}{c|}{\underline{10.009 $\pm$ ${\scriptstyle 0.114}$}} & \multicolumn{1}{c|}{\underline{19.734 $\pm$ ${\scriptstyle 0.228}$}} & \multicolumn{1}{c|}{1117.9 $\pm$ ${\scriptstyle 28.6}$} & 0.012 $\pm$ ${\scriptstyle 0.001}$ \\
    \multicolumn{1}{|l|}{} & \multicolumn{1}{l|}{BMA} & \multicolumn{1}{c|}{0.741 $\pm$ ${\scriptstyle 0.009}$} & \multicolumn{1}{c|}{7.934 $\pm$ ${\scriptstyle 0.084}$} & \multicolumn{1}{c|}{15.642 $\pm$ ${\scriptstyle 0.172}$} & \multicolumn{1}{c|}{30.258 $\pm$ ${\scriptstyle 0.344}$} & \multicolumn{1}{c|}{1998.1 $\pm$ ${\scriptstyle 42.6}$} & 0.062 $\pm$ ${\scriptstyle 0.002}$ \\
    \multicolumn{1}{|l|}{} & \multicolumn{1}{l|}{Linear Reg} & \multicolumn{1}{c|}{0.503 $\pm$ ${\scriptstyle 0.004}$} & \multicolumn{1}{c|}{5.741 $\pm$ ${\scriptstyle 0.068}$} & \multicolumn{1}{c|}{11.236 $\pm$ ${\scriptstyle 0.128}$} & \multicolumn{1}{c|}{21.945 $\pm$ ${\scriptstyle 0.256}$} & \multicolumn{1}{c|}{1296.9 $\pm$ ${\scriptstyle 31.2}$} & 0.031 $\pm$ ${\scriptstyle 0.001}$ \\
    \multicolumn{1}{|l|}{} & \multicolumn{1}{l|}{HGB} & \multicolumn{1}{c|}{0.501 $\pm$ ${\scriptstyle 0.004}$} & \multicolumn{1}{c|}{5.824 $\pm$ ${\scriptstyle 0.062}$} & \multicolumn{1}{c|}{11.446 $\pm$ ${\scriptstyle 0.118}$} & \multicolumn{1}{c|}{22.458 $\pm$ ${\scriptstyle 0.234}$} & \multicolumn{1}{c|}{1531.9 $\pm$ ${\scriptstyle 34.2}$} & 0.004 $\pm$ ${\scriptstyle 0.001}$ \\
    \multicolumn{1}{|l|}{} & \multicolumn{1}{l|}{RFR} & \multicolumn{1}{c|}{0.595 $\pm$ ${\scriptstyle 0.002}$} & \multicolumn{1}{c|}{6.919 $\pm$ ${\scriptstyle 0.018}$} & \multicolumn{1}{c|}{13.620 $\pm$ ${\scriptstyle 0.036}$} & \multicolumn{1}{c|}{26.657 $\pm$ ${\scriptstyle 0.075}$} & \multicolumn{1}{c|}{1369.1 $\pm$ ${\scriptstyle 37.9}$} & 0.070 $\pm$ ${\scriptstyle 0.003}$ \\ \cmidrule(lr){2-8} 
    \multicolumn{1}{|l|}{} & \multicolumn{1}{l|}{\textbf{MP-MoE (Ours)}} & \multicolumn{1}{c|}{\textbf{0.357 $\pm$ ${\scriptstyle 0.007}$}} & \multicolumn{1}{c|}{\textbf{3.639 $\pm$ ${\scriptstyle 0.126}$}} & \multicolumn{1}{c|}{\textbf{7.091 $\pm$ ${\scriptstyle 0.250}$}} & \multicolumn{1}{c|}{\textbf{13.764 $\pm$ ${\scriptstyle 0.492}$}} & \multicolumn{1}{c|}{\textbf{805.0 $\pm$ ${\scriptstyle 15.4}$}} & \textbf{0.216 $\pm$ ${\scriptstyle 0.017}$} \\ \midrule
    \multicolumn{8}{|c|}{\textbf{Panel B: Song Chay Basin}} \\ \midrule
    \multicolumn{1}{|l|}{\multirow{6}{*}{\begin{tabular}[c]{@{}l@{}}Physics-Based\\ Experts\end{tabular}}} & \multicolumn{1}{l|}{WRF84H} & \multicolumn{1}{c|}{0.585} & \multicolumn{1}{c|}{6.209} & \multicolumn{1}{c|}{12.037} & \multicolumn{1}{c|}{22.463} & \multicolumn{1}{c|}{3879.2} & \underline{0.083} \\
    \multicolumn{1}{|l|}{} & \multicolumn{1}{l|}{COMS} & \multicolumn{1}{c|}{0.500} & \multicolumn{1}{c|}{5.442} & \multicolumn{1}{c|}{10.440} & \multicolumn{1}{c|}{19.534} & \multicolumn{1}{c|}{2992.9} & 0.068 \\
    \multicolumn{1}{|l|}{} & \multicolumn{1}{l|}{GFS} & \multicolumn{1}{c|}{0.459} & \multicolumn{1}{c|}{5.077} & \multicolumn{1}{c|}{9.884} & \multicolumn{1}{c|}{18.824} & \multicolumn{1}{c|}{2780.2} & 0.044 \\
    \multicolumn{1}{|l|}{} & \multicolumn{1}{l|}{ICON} & \multicolumn{1}{c|}{\underline{0.405}} & \multicolumn{1}{c|}{\underline{4.662}} & \multicolumn{1}{c|}{\underline{9.159}} & \multicolumn{1}{c|}{\underline{17.707}} & \multicolumn{1}{c|}{\underline{2692.5}} & 0.025 \\
    \multicolumn{1}{|l|}{} & \multicolumn{1}{l|}{MIT\_D01} & \multicolumn{1}{c|}{0.666} & \multicolumn{1}{c|}{7.500} & \multicolumn{1}{c|}{14.537} & \multicolumn{1}{c|}{27.848} & \multicolumn{1}{c|}{3729.4} & 0.056 \\
    \multicolumn{1}{|l|}{} & \multicolumn{1}{l|}{MIT} & \multicolumn{1}{c|}{0.703} & \multicolumn{1}{c|}{7.825} & \multicolumn{1}{c|}{15.192} & \multicolumn{1}{c|}{28.630} & \multicolumn{1}{c|}{4037.7} & 0.071 \\ \midrule
    \multicolumn{1}{|l|}{\multirow{8}{*}{\begin{tabular}[c]{@{}l@{}}Learning-based\\ Models\end{tabular}}} & \multicolumn{1}{l|}{Ensemble} & \multicolumn{1}{c|}{1.105 $\pm$ ${\scriptstyle 0.021}$} & \multicolumn{1}{c|}{12.800 $\pm$ ${\scriptstyle 0.270}$} & \multicolumn{1}{c|}{25.016 $\pm$ ${\scriptstyle 0.543}$} & \multicolumn{1}{c|}{47.882 $\pm$ ${\scriptstyle 1.093}$} & \multicolumn{1}{c|}{5474.3 $\pm$ ${\scriptstyle 137.9}$} & 0.073 $\pm$ ${\scriptstyle 0.003}$ \\
    \multicolumn{1}{|l|}{} & \multicolumn{1}{l|}{XGBoost} & \multicolumn{1}{c|}{1.866 $\pm$ ${\scriptstyle 0.018}$} & \multicolumn{1}{c|}{20.359 $\pm$ ${\scriptstyle 0.245}$} & \multicolumn{1}{c|}{40.000 $\pm$ ${\scriptstyle 0.480}$} & \multicolumn{1}{c|}{77.932 $\pm$ ${\scriptstyle 0.924}$} & \multicolumn{1}{c|}{10052.2 $\pm$ ${\scriptstyle 201.0}$} & 0.074 $\pm$ ${\scriptstyle 0.002}$ \\
    \multicolumn{1}{|l|}{} & \multicolumn{1}{l|}{LSTM} & \multicolumn{1}{c|}{1.839 $\pm$ ${\scriptstyle 0.126}$} & \multicolumn{1}{c|}{20.019 $\pm$ ${\scriptstyle 1.563}$} & \multicolumn{1}{c|}{39.182 $\pm$ ${\scriptstyle 3.128}$} & \multicolumn{1}{c|}{76.215 $\pm$ ${\scriptstyle 6.273}$} & \multicolumn{1}{c|}{9840.5 $\pm$ ${\scriptstyle 577.3}$} & 0.079 $\pm$ ${\scriptstyle 0.008}$ \\
    \multicolumn{1}{|l|}{} & \multicolumn{1}{l|}{QRA} & \multicolumn{1}{c|}{0.528 $\pm$ ${\scriptstyle 0.006}$} & \multicolumn{1}{c|}{6.116 $\pm$ ${\scriptstyle 0.078}$} & \multicolumn{1}{c|}{11.920 $\pm$ ${\scriptstyle 0.144}$} & \multicolumn{1}{c|}{22.723 $\pm$ ${\scriptstyle 0.264}$} & \multicolumn{1}{c|}{2852.5 $\pm$ ${\scriptstyle 56.4}$} & 0.025 $\pm$ ${\scriptstyle 0.001}$ \\
    \multicolumn{1}{|l|}{} & \multicolumn{1}{l|}{BMA} & \multicolumn{1}{c|}{0.522 $\pm$ ${\scriptstyle 0.006}$} & \multicolumn{1}{c|}{5.843 $\pm$ ${\scriptstyle 0.072}$} & \multicolumn{1}{c|}{11.416 $\pm$ ${\scriptstyle 0.142}$} & \multicolumn{1}{c|}{21.914 $\pm$ ${\scriptstyle 0.268}$} & \multicolumn{1}{c|}{3274.2 $\pm$ ${\scriptstyle 64.8}$} & 0.017 $\pm$ ${\scriptstyle 0.001}$ \\
    \multicolumn{1}{|l|}{} & \multicolumn{1}{l|}{Linear Reg} & \multicolumn{1}{c|}{1.034 $\pm$ ${\scriptstyle 0.012}$} & \multicolumn{1}{c|}{12.023 $\pm$ ${\scriptstyle 0.144}$} & \multicolumn{1}{c|}{23.436 $\pm$ ${\scriptstyle 0.282}$} & \multicolumn{1}{c|}{44.651 $\pm$ ${\scriptstyle 0.536}$} & \multicolumn{1}{c|}{4417.1 $\pm$ ${\scriptstyle 88.4}$} & 0.034 $\pm$ ${\scriptstyle 0.001}$ \\
    \multicolumn{1}{|l|}{} & \multicolumn{1}{l|}{HGB} & \multicolumn{1}{c|}{0.927 $\pm$ ${\scriptstyle 0.057}$} & \multicolumn{1}{c|}{10.719 $\pm$ ${\scriptstyle 0.665}$} & \multicolumn{1}{c|}{20.891 $\pm$ ${\scriptstyle 1.316}$} & \multicolumn{1}{c|}{39.796 $\pm$ ${\scriptstyle 2.650}$} & \multicolumn{1}{c|}{5570.7 $\pm$ ${\scriptstyle 180.0}$} & 0.062 $\pm$ ${\scriptstyle 0.008}$ \\
    \multicolumn{1}{|l|}{} & \multicolumn{1}{l|}{RFR} & \multicolumn{1}{c|}{1.750 $\pm$ ${\scriptstyle 0.022}$} & \multicolumn{1}{c|}{20.272 $\pm$ ${\scriptstyle 0.262}$} & \multicolumn{1}{c|}{39.935 $\pm$ ${\scriptstyle 0.525}$} & \multicolumn{1}{c|}{78.289 $\pm$ ${\scriptstyle 1.069}$} & \multicolumn{1}{c|}{7711.3 $\pm$ ${\scriptstyle 161.1}$} & 0.074 $\pm$ ${\scriptstyle 0.001}$ \\ \cmidrule(lr){2-8} 
    \multicolumn{1}{|l|}{} & \multicolumn{1}{l|}{\textbf{MP-MoE (Ours)}} & \multicolumn{1}{c|}{\textbf{0.282 $\pm$ ${\scriptstyle 0.005}$}} & \multicolumn{1}{c|}{\textbf{3.106 $\pm$ ${\scriptstyle 0.064}$}} & \multicolumn{1}{c|}{\textbf{6.100 $\pm$ ${\scriptstyle 0.125}$}} & \multicolumn{1}{c|}{\textbf{11.792 $\pm$ ${\scriptstyle 0.245}$}} & \multicolumn{1}{c|}{\textbf{1346.3 $\pm$ ${\scriptstyle 41.4}$}} & \textbf{0.174 $\pm$ ${\scriptstyle 0.010}$} \\ \midrule
    \end{tabular}}
\end{table*}
    
    \noindent\textbf{Baselines.}~To evaluate the efficacy of the MP-MoE framework, we consider two distinct categories of benchmarks: (i) a wide range of learning-based models, including regression-based ensembles and recurrent neural networks, and (ii) operational NWP experts. This setup is designed to isolate the specific contribution of our structural-aware gating mechanism in improving forecast fidelity over both standard point-wise ensemble techniques and purely physical models. By balancing subsequence matching with intensity calibration, we assess the framework's ability to bridge the gap between statistical post-processing and localized structural accuracy. 
    
    \begin{figure*}[t]
    \centering
    \begin{subfigure}[t]{0.49\textwidth}
        \centering
        \includegraphics[width=\linewidth]{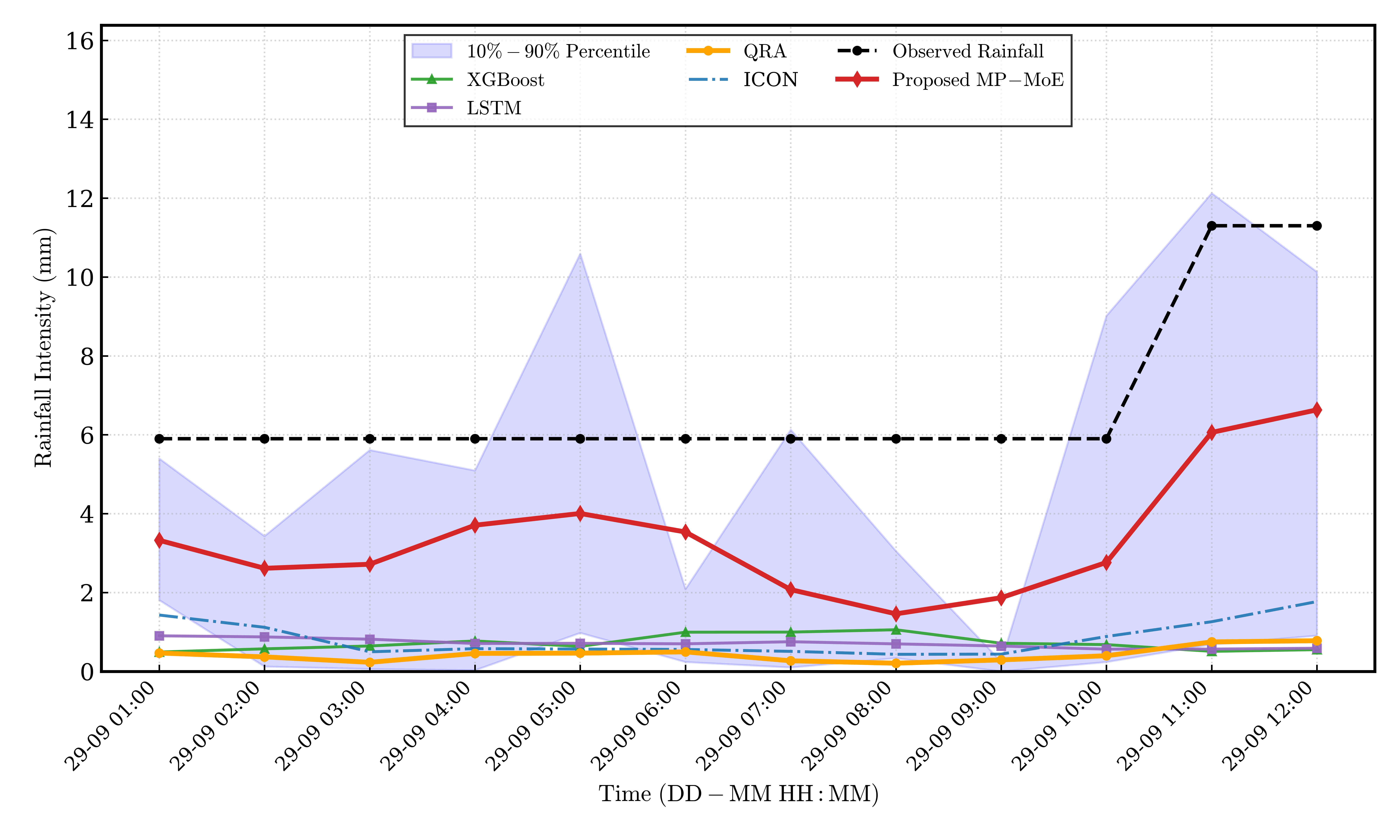}
        \caption{Ban Nhung Basin} 
        \label{fig:BanNhung}
    \end{subfigure}
    \hfill
    \begin{subfigure}[t]{0.49\textwidth}
        \centering
        \includegraphics[width=\linewidth]{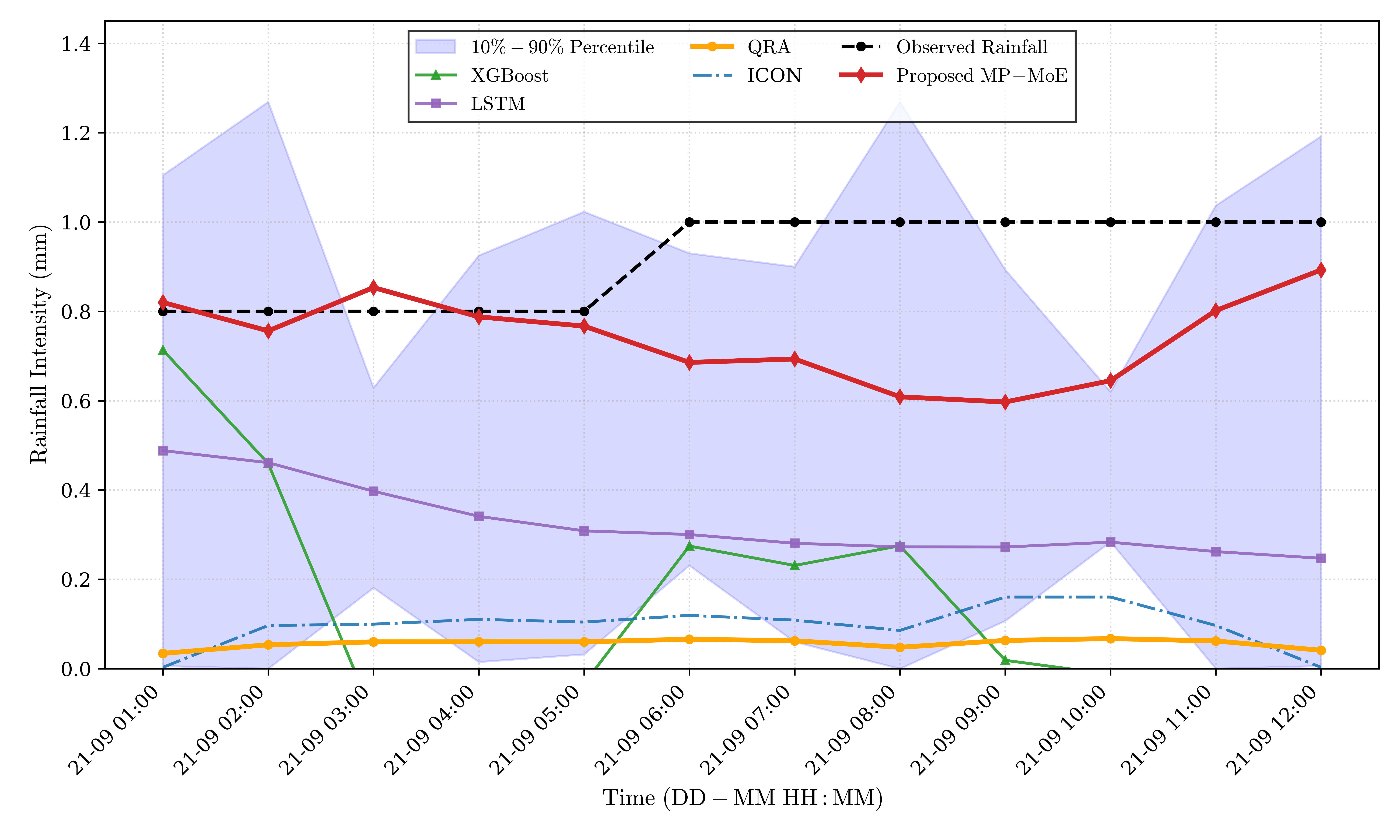}
        \caption{Song Chay Basin} 
        \label{fig:SongChay}
    \end{subfigure}
    \caption{Comparative visualization of forecast trajectories. The proposed MP-MoE ($\lambda=0.6$, solid red line) effectively captures high-intensity peaks and rapid onsets compared to the ground truth (black dashed line), whereas traditional baselines exhibit significant peak-shaving effects in the Ban Nhung and Song Chay basins.}
    \label{fig:CompareBasins}
    \end{figure*}
    
    \noindent\textbf{Evaluation Metrics.}~Model performance is assessed using three complementary metrics: (i) \texttt{MAE} for 1-hour and accumulated rainfall (12h, 24h, 48h) to evaluate volumetric error; (ii) \texttt{DTW} to quantify temporal shape alignment; and (iii) \texttt{Mean Critical Success Index} (CSI-M). Specifically, CSI-M averages CSI scores across thresholds $r \in \{1, 3, 5 \text{ mm}\}$, where continuous values are binarized ($\hat{y} \ge r$) to calculate Hits, Misses, and False Alarms. This multi-threshold approach captures the model's capability in detecting events ranging from light to high intensities.
    
    \noindent\textbf{Implementation Details.}~Our framework is implemented in PyTorch and optimized using the Adam optimizer with a learning rate of 0.003 and a batch size of 64. The Gating Network employs a lightweight four-layer MLP to dynamically compute expert weights. To balance structural fidelity with computational overhead, we set the subsequence length $m$ and maximum permissible temporal shift $\Delta$ are both set to 3 hours. A trade-off parameter $\lambda$ is further introduced to regulate the hybrid loss function. Robustness is verified through five independent runs across seeds $\{0, 1, 42, 2024, 2025\}$, with results reported as mean $\pm$ standard deviation (STD) for stochastic models.
    \subsection{Comparative Performance}
    \label{sec:Comparative_Performance}
    \noindent\textbf{Quantitative Performance.}~Table \ref{tab:results_table} shows that while certain baselines excel in specific regions, MP-MoE maintains consistent accuracy across diverse hydrological settings. In the Ban Nhung basin, the QRA model is a competitive statistical baseline with a 1-hour MAE of 0.440 mm, whereas the physical ICON expert provides a more reliable benchmark in the Song Chay basin with an MAE of 0.405 mm. MP-MoE outperforms both, achieving 1-hour MAEs of 0.357 mm and 0.282 mm for Ban Nhung and Song Chay, respectively.

    This consistency extends to structural metrics, where MP-MoE yields the minimum DTW distances in both basins, reaching 805.0 in Ban Nhung compared to 997.0 from the LSTM model. The framework also improves rainfall event detection, with a CSI-M of 0.174 in Song Chay, nearly double the scores of the WRF84H and LSTM baselines. For 48-hour accumulated rainfall, MP-MoE reduces systematic biases by lowering the Ban Nhung MAE to 13.764 mm, a substantial gain over the 24.086 mm from the top deterministic expert. These results suggest that MP guidance helps the gating network effectively adapt to varying meteorological regimes.
   
    \noindent\textbf{Structural preservation in extreme transition regimes.}~Fig.~\ref{fig:BanNhung} illustrates a sharp rainfall surge in the Ban Nhung basin from 6 mm/h to nearly 12 mm/h. While all baselines fail to react and remain near the zero-axis, MP-MoE reflects the storm morphology by exhibiting a clear upward trajectory. Specifically, the proposed model reaches an intensity of approximately 6.7 mm/h, notably outperforming the flattened predictions of LSTM and ICON. Although the absolute peak remains underestimated, this ability to capture the rising trend demonstrates that the MP objective provides a superior gradient signal for structural alignment during extreme transitions.

    \noindent\textbf{Consistency in sustained rainfall.}~The Song Chay basin (Fig.~\ref{fig:SongChay}) illustrates a sustained rainfall scenario, where observed intensity remains near 1.0 mm/h. In this setting, baselines such as LSTM and XGBoost exhibit a dissipation bias, with predicted values rapidly decaying toward zero despite ongoing precipitation. By comparison, MP-MoE preserves temporal persistence by maintaining levels around 0.8 mm/h. This demonstrates that our model effectively retains continuous patterns and mitigates the premature attenuation typical of these data-driven benchmarks.
    
\begin{figure}[t!]
        \centering
        \includegraphics[width=\columnwidth]{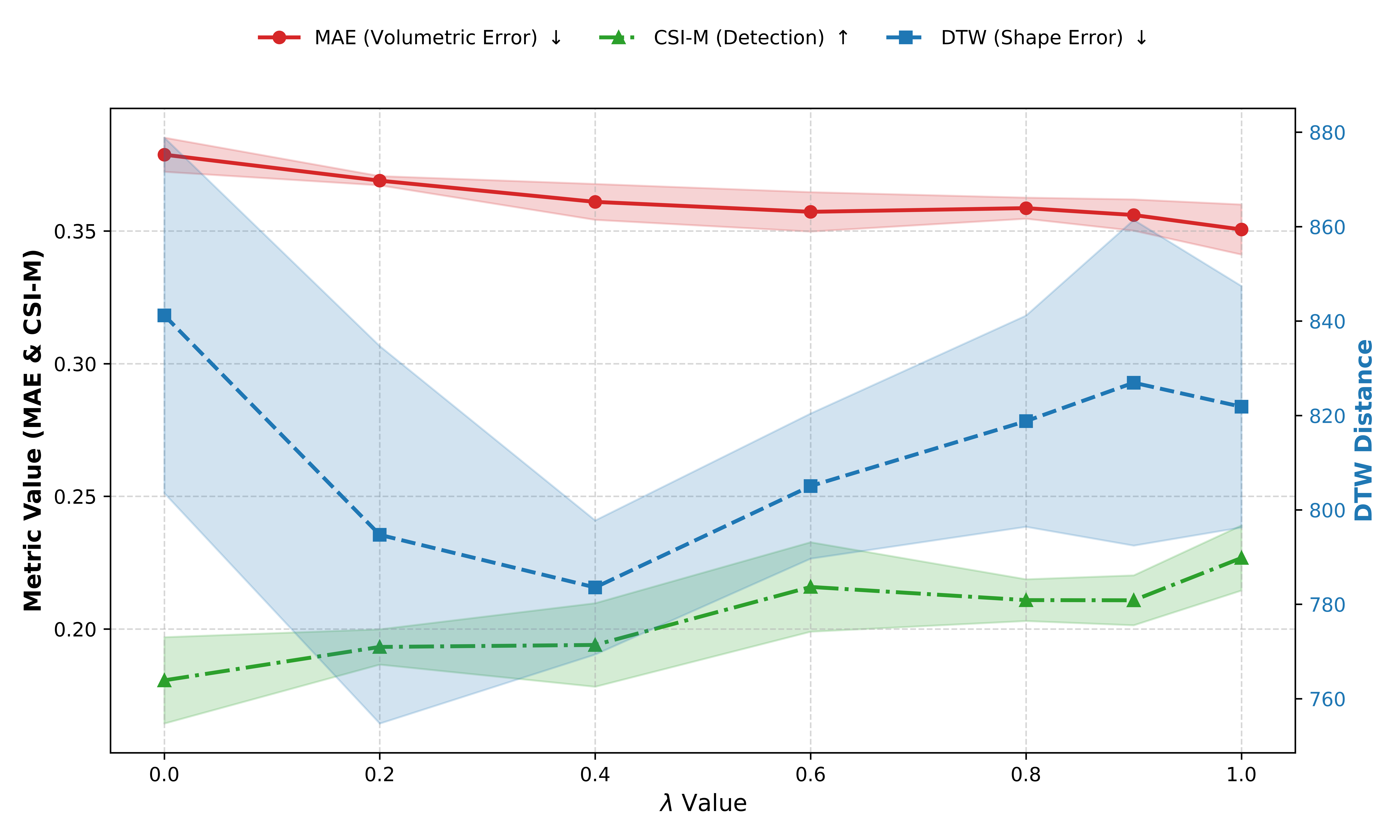}
        \caption{Sensitivity analysis (at Ban Nhung) of the hyperparameter $\lambda$ which regulates the trade-off between the intensity-based MSE loss and the Modified MP loss. The dual-axis chart shows MAE and CSI-M on the left axis, and DTW distance on the right axis.}
        \label{fig:lambda_balance}
    \end{figure}

    \noindent\textbf{Parameter sensitivity analysis.}~The hyperparameter $\lambda$ in \eqref{eq:loss_total} controls the balance between point-wise intensity calibration and structural alignment. As illustrated in Fig.~\ref{fig:lambda_balance}, the choice of $\lambda$ dictates the model priority across different forecast requirements. Specifically, if the primary goal is to minimize hourly intensity errors, a lower $\lambda$ value focusing on MSE-based calibration is more suitable, though this often leads to the peak-shaving effect and poor event detection. Conversely, if the objective is to maximize categorical detection and shape similarity, a higher $\lambda$ value approaching 1.0 becomes preferable as it prioritizes subsequence matching over rigid point-wise alignment. We identify $\lambda=0.6$ as the most harmonious configuration for operational reliability. In this setting, the model achieves a high CSI-M of 0.216 while significantly reducing the DTW distance, thereby maintaining necessary anchor-to-physical-magnitude constraints while preventing the non-physical temporal warping associated with exclusive subsequence optimization. Consequently, this intermediate balance ensures that the gating network selects experts with both morphological fidelity and magnitude consistency across varying rainfall regimes.
\subsection{Ablation Study}
    \begin{table}[!h]
    \centering
    \caption{Ablation study results focusing on structural fidelity DTW $\downarrow$ across two river basins. Results are reported as Mean $\pm$ STD over five independent runs.}
    \label{tab:ablation_dtw}
    \resizebox{\columnwidth}{!}{
    \begin{tabular}{l|ccc}
        \toprule
        \textbf{Basin / Configuration} & \textbf{MP-MoE (Full)} & \textbf{w/o MP Loss} & \textbf{w/o MSE Loss} \\ \midrule
        \midrule
        Ban Nhung (DTW $\downarrow$) & $\mathbf{805.0} \pm {\scriptstyle 15.4}$ & $841.2 \pm {\scriptstyle 37.7}$ & $821.8 \pm {\scriptstyle 25.6}$ \\
        \midrule
        Song Chay (DTW $\downarrow$) & $\mathbf{1346.3} \pm {\scriptstyle 41.4}$ & $1350.4 \pm {\scriptstyle 34.7}$ & $1782.2 \pm {\scriptstyle 48.0}$ \\
        \bottomrule
    \end{tabular}
    }
\end{table}
    Table~\ref{tab:ablation_dtw} evaluates the contributions of MSE and MP components to structural fidelity. Omitting either term degrades performance, confirming the necessity of a hybrid objective. At Ban Nhung, excluding MP loss increases DTW to 841.2, showing that point-wise metrics alone struggle with temporal shifts. Similarly, Song Chay exhibits a substantial DTW increase to 1782.2 without MSE loss, as subsequence matching without intensity calibration leads to volumetric distortions. By integrating both objectives, MP-MoE achieves optimal stability, minimizing DTW to 805.0 and 1346.3. These findings validate that structural guidance prevents temporal warping while maintaining magnitude consistency.
    

%% file: secs/Limitations.tex
\section{Limitations and Future Work}
    While the MP-MoE framework significantly advances structural forecast accuracy, it lays a foundation with several directions for future development. The current implementation focuses on 1D time-series modeling at the basin scale, a choice that prioritizes local structural fidelity over broader spatiotemporal field correlations. Consequently, while the gating network effectively routes trust among NWP experts, its performance is bounded by the diversity of the available expert pool. Furthermore, the use of fixed hyperparameters, such as maximum permissible temporal shift $\Delta$ and the loss-balancing coefficient $\lambda$, represents an initial baseline that could be further refined through adaptive tuning to better capture the volatility of tropical monsoon regimes. Future research will focus on extending the framework to the spatiotemporal domain by incorporating graph-based dependencies, ensuring regional consistency across neighboring catchments. Additionally, exploring online learning mechanisms will allow the Gating Network to autonomously adapt to seasonal climate shifts and evolving atmospheric patterns, further enhancing the system's operational resilience.

%% file: secs/Conclusion.tex
\section{Conclusion}
    This paper presented MP-MoE, a framework designed to address temporal misalignment in precipitation forecasting. By shifting the optimization paradigm to structural similarity via the hybrid loss objective, the model effectively mitigates the double penalty and peak-shaving biases. Evaluations in Vietnam demonstrate that MP-MoE consistently outperforms operational NWP models and learning-based baselines, achieving substantial reductions in DTW and improvements in CSI-M. These results validate the framework's reliability for operational flood early warning and disaster management.

\section*{Acknowledgment}
    The authors gratefully acknowledge WEATHERPLUS Solution Joint Stock Company for providing the meteorological datasets used in this study, and AITHINGS Co. Ltd. for their support in conducting this research.